\documentclass{article}

    \PassOptionsToPackage{numbers, compress}{natbib}
\usepackage[final]{neurips_2024}

\usepackage[utf8]{inputenc} 
\usepackage[T1]{fontenc}    
\usepackage{hyperref}       
\usepackage{url}            
\usepackage{booktabs}       
\usepackage{amsfonts}       
\usepackage{nicefrac}       
\usepackage{microtype}      
\usepackage{xcolor}         

\usepackage[utf8]{inputenc}
\usepackage{amsmath} 
\usepackage{amssymb} 
\usepackage{amsfonts} 
\usepackage{graphicx} 

\usepackage{algorithm}
\usepackage{algpseudocode}

\usepackage{booktabs}

\usepackage{marvosym}
\usepackage{authblk} 

\title{Cross-Modality Perturbation Synergy Attack for Person Re-identification}

\author[1]{Yunpeng Gong}
\author[2]{Zhun Zhong}
\author[1]{Yansong Qu}
\author[1]{Zhiming Luo}
\author[1]{Rongrong Ji}
\author[1]{Min Jiang\textsuperscript{*}}

\affil[1]{School of Informatics, Xiamen University}
\affil[2]{School of Computer Science and Information Engineering, Hefei University of Technology}

\date{}  

\begin{document}

\maketitle

\renewcommand{\thefootnote}{*}
\footnotetext{Min Jiang and Yunpeng Gong are with the Department of Artificial
	Intelligence, Key Laboratory of Multimedia Trusted Perception and Efficient Computing, Ministry of Education of China, School of Informatics, Key Laboratory of Digital Protection and Intelligent Processing of Intangible CulturalHeritage of Fujian and Taiwan, Ministry of Culture and Tourism, Xiamen University, Xiamen 361005, Fujian, P.R. China (e-mail: minjiang@xmu.edu.cn; gongyunpeng@stu.xmu.edu.cn or fmonkey625@gmail.com).\\
	\hspace*{2em} Corresponding author: Min Jiang}

\begin{abstract}
In recent years, there has been significant research focusing on addressing security concerns in single-modal person re-identification (ReID) systems that are based on RGB images. However, the safety of cross-modality scenarios, which are more commonly encountered in practical applications involving images captured by infrared cameras, has not received adequate attention. The main challenge in cross-modality ReID lies in effectively dealing with visual differences between different modalities. For instance, infrared images are typically grayscale, unlike visible images that contain color information. Existing attack methods have primarily focused on the characteristics of the visible image modality, overlooking the features of other modalities and the variations in data distribution among different modalities. This oversight can potentially undermine the effectiveness of these methods in image retrieval across diverse modalities. This study represents the first exploration into the security of cross-modality ReID models and proposes a universal perturbation attack specifically designed for cross-modality ReID. This attack optimizes perturbations by leveraging gradients from diverse modality data, thereby disrupting the discriminator and reinforcing the differences between modalities. We conducted experiments on three widely used cross-modality datasets, namely RegDB, SYSU, and LLCM. The results not only demonstrate the effectiveness of our method but also provide insights for future improvements in the robustness of cross-modality ReID systems.
\end{abstract}

\section{Introduction}

With the rapid advancement of surveillance technology, person re-identification (ReID)~\cite{yang2024shallow,zuo2024ufinebench,gong2024exploring,shi2023dual} has emerged as a pivotal component in the realm of security, garnering escalating attention. ReID constitutes a fundamental task in computer vision~\cite{ma2023towards,wu20243d}, aiming to precisely identify the same individual across diverse locations and time points by analyzing pedestrian images captured through surveillance cameras~\cite{9018132}. The challenges inherent in this task encompass factors such as changes in viewpoint, lighting conditions~\cite{gong2024beyond,gong2021eliminate}, occlusion~\cite{tan2022dynamic,tan2024occluded}, and pose variations, culminating in significant appearance variations of the same individual across distinct camera views~\cite{Zhong_2018_CVPR}.

\begin{figure}[t]
	\setlength{\abovecaptionskip}{-0.1cm}
	\setlength{\belowcaptionskip}{-0.3cm}   
	\centering
	\includegraphics[width=0.8\linewidth]{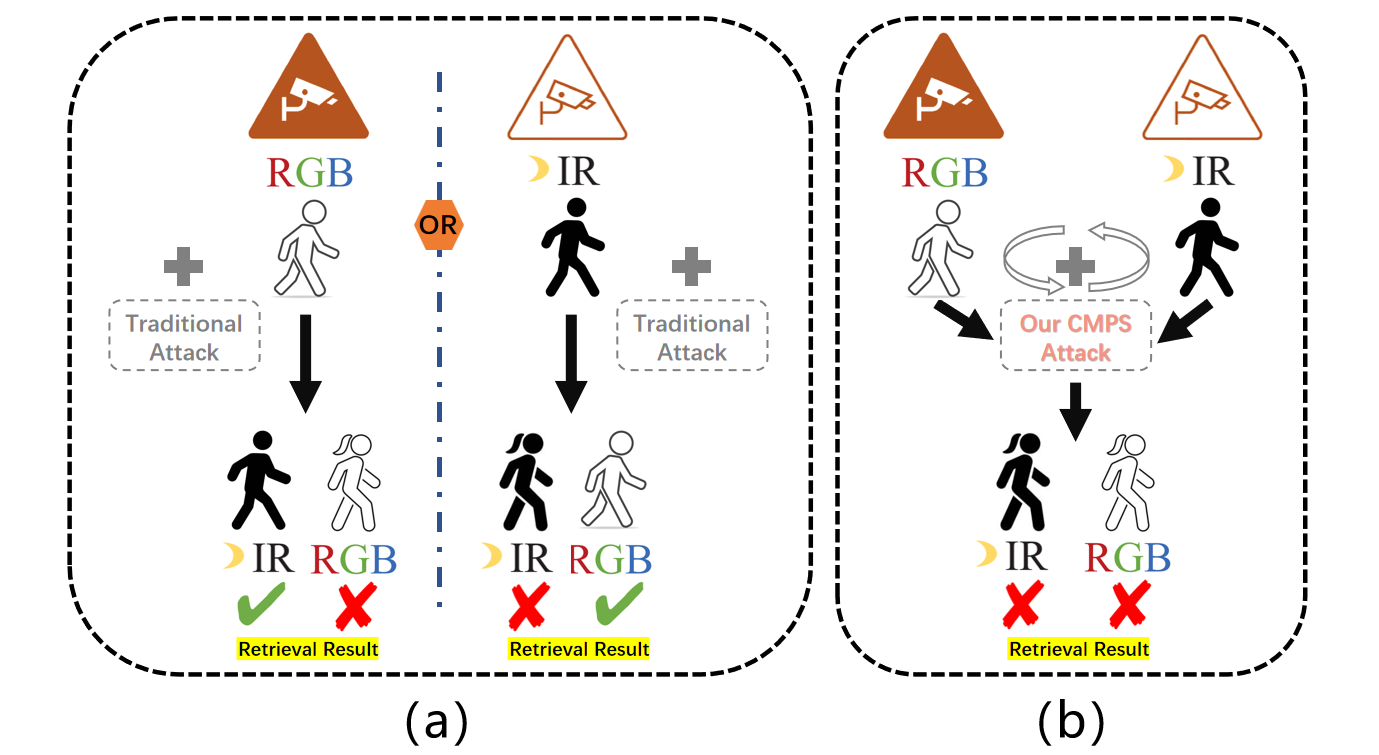}
	\caption{Comparison between traditional and proposed methods: Fig.(a) illustrates traditional attack methods (e.g., FGSM~\cite{fgsm}, PGD~\cite{madry2018towards}), which are primarily designed for single-modal tasks and lack mechanisms to associate multiple modalities, making them ineffective in simultaneously misleading retrieval results across different modalities. Fig.(b) illustrates the proposed method, which employs an intrinsic mechanism to effectively associate different modalities, thereby misleading retrieval results across multiple modalities simultaneously.}
\label{pic1}
\end{figure}

In traditional ReID, where samples are image-based, the conventional methodology centers on matching visible to visible (RGB to RGB) data. However, when dealing with diverse scenarios and conditions, especially involving multiple image modalities such as RGB and infrared images, the system needs to intricately handle the differences in images from different modalities~\cite{zhang2023mrcn,shi2024learningcommonalitydivergencevariety,Yang_2023_ICCV}. This is essential to ensure that the system exhibits better robustness across different modalities. Hence, cross-modality ReID is considered more challenging due to the need for addressing these modality differences~\cite{ye2021deep,RPLNR}.

Cross-modal ReID~\cite{dpis,zhang2023diverse,ye2021deep,eccv20ddag} plays a crucial role in significantly expanding the applicability of traditional ReID methods, focusing on addressing complex matching issues between different image modalities. In practical surveillance systems, the simultaneous use of multiple sensors, such as RGB cameras and infrared cameras, is a common scenario. This task requires innovative solutions to effectively bridge the differences between various modalities, ensuring robust and accurate re-identification of pedestrians in heterogeneous sensor outputs. 

\begin{figure*}[]
	\centering
	\setlength{\abovecaptionskip}{-0cm}
	\setlength{\belowcaptionskip}{-0.2cm}   
	\includegraphics[width=0.9\linewidth]{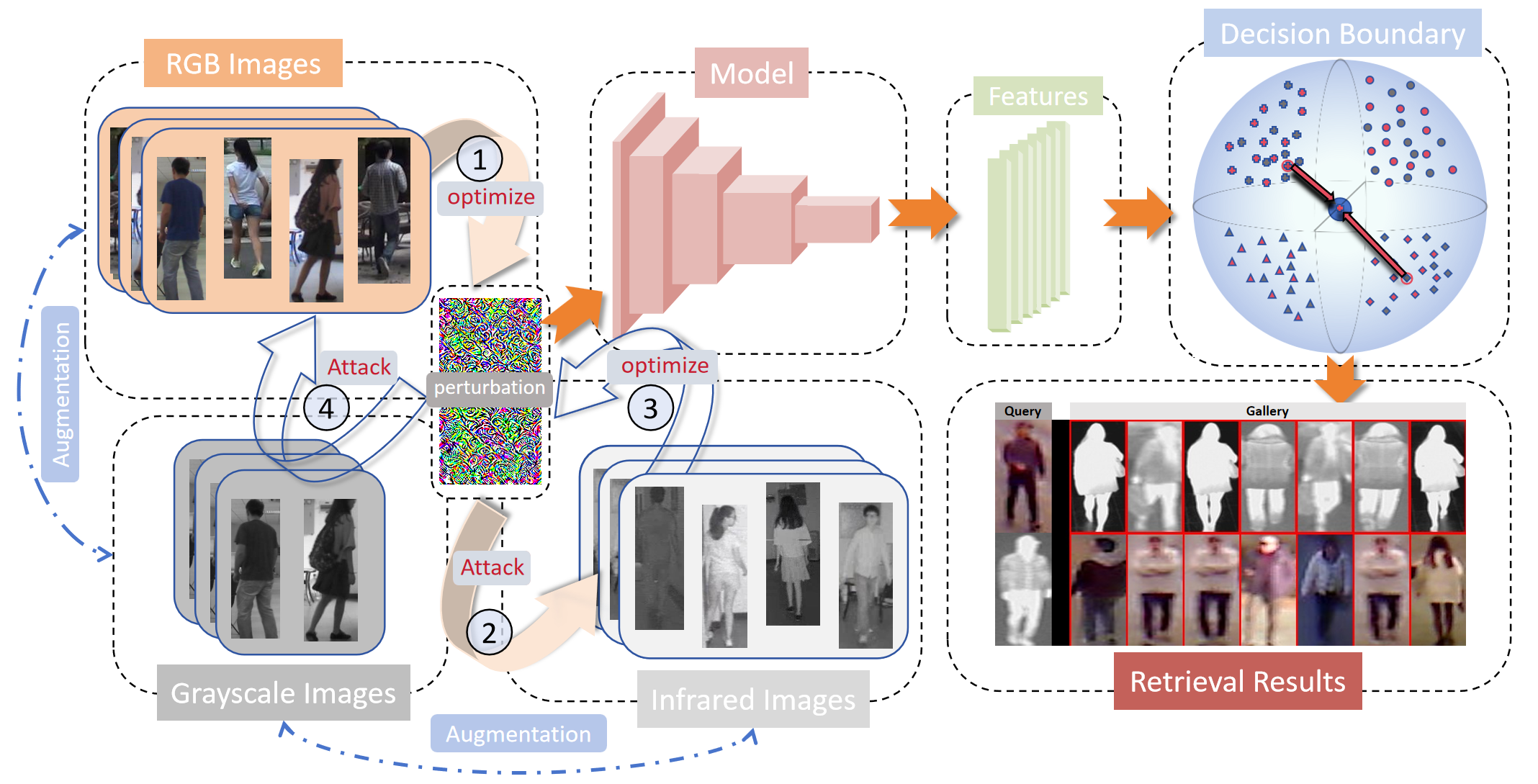}
	\caption{Illustration of the CMPS attack framework. We generate homogeneous grayscale images through random grayscale transformations to reduce the differences between modalities, aiding in the learning of a universal perturbation. The process is as follows: first, the gradient from one modality is used to optimize the universal perturbation, which is then applied to another modality's images to generate adversarial samples for attacks. The new modality's gradient is then used to further optimize the perturbation and attack the next modality. By aggregating feature gradients from different modalities, we iteratively learn a universal perturbation, pushing samples toward a common region in the manifold. The manifold is represented as a sphere, with identical shapes but different colors representing the same person's features across modalities. This method captures shared knowledge between modalities, enabling more effective learning of cross-modal universal perturbations.}
	\label{pipline}
\end{figure*}

Currently, most research on the security of ReID focuses on single-modality systems based on RGB images~\cite{zheng2023u,yang2023towards,gong2022person,bai2020adversarial,bouniot2020vulnerability,wang2020transferable}, while the security of cross-modality ReID systems has received insufficient attention. The challenge in cross-modality attacks arises from significant visual differences among different modality inputs, requiring attackers to effectively capture shared features from each modality for perturbation implementation. However, as shown in Fig.~\ref{pic1}, existing attack methods in cross-modal scenarios require optimizing perturbations separately for each modality, lacking an intrinsic mechanism to capture shared knowledge between different modalities, which limits the success rate of the attacks. To address this issue, we propose a synergistic optimization method combined with triplet loss, utilizing information from different modalities to optimize the universal perturbation. This method pushes the features of different samples into a common sub-region that affects the model's accuracy, as shown in Fig.~\ref{pipline}.

Specifically, we propose the Cross-Modality Perturbation Synergy (CMPS) method, a universal perturbation approach designed specifically for cross-modality ReID systems. This method simultaneously leverages gradient information from multiple modalities to jointly optimize universal perturbations across visible and infrared images. CMPS incorporates cross-modality triplet loss to ensure feature consistency across different modalities, enhancing the generality of the perturbation. During the synergistic optimization process, CMPS iteratively updates gradients from various modalities within a unified optimization framework, effectively capturing and utilizing shared features across modalities. To further reduce visual differences between modalities, we introduce cross-modality attack augmentation, converting images into grayscale to standardize their visual representation and facilitate the learning of modality-agnostic perturbations. As a result, these universal perturbations push the features of different samples toward a common region in the feature space, significantly diminishing the model's ability to accurately distinguish identities in cross-modality scenarios, thereby successfully deceiving the model.

In our experiments on widely utilized cross-modality ReID datasets, including RegDB~\cite{nguyen2017person}, SYSU~\cite{wu2017rgb} and LLCM~\cite{zhang2023diverse}, we not only showcase the effectiveness of our proposed method but also provide insights for fortifying the robustness of cross-modality ReID systems in the future. This research contributes by bridging gaps in current studies and introducing novel perspectives to study the security challenges in cross-modality ReID systems.

The main contributions of our work can be summarized as:

$\bullet$ To the best of our knowledge, our work is the first to investigate vulnerabilities in cross-modality ReID models. By explicitly incorporating cross-modality constraints into the synergistic optimization process, we enhance the universality of the learned cross-modality perturbations. Additionally, we provide mathematical analysis to demonstrate the superiority of our proposed method over traditional approaches.

$\bullet$ We propose a cross-modality attack augmentation method, utilizing random grayscale transformations to narrow the gap between different modalities, aiding our cross-modality perturbation synergy attack in better capturing shared features across modalities.

$\bullet$ Extensive experiments conducted on three widely used cross-modality ReID benchmarks demonstrate the effectiveness of our proposed cross-modality attack. Our method exhibits good transferability even when attacking different models. The code will be available at \url{https://github.com/finger-monkey/cmps__attack}.

\section{Related Works}

\textbf{Adversarial Attack.}
Adversarial attacks are a technique involving the clever design of small input perturbations with the aim of deceiving machine learning models, leading them to produce misleading outputs. This form of attack is not confined to the image domain but extends to models in various fields, including speech~\cite{Wang2023Towards} and text~\cite{Wang2021Towards,ma2022xclip,wu2024controlmllm}. Typically, the goal of adversarial attacks is to tweak input data in a way that causes the model to make erroneous predictions when handling these subtly modified samples~\cite{fgsm,carlini2017towards,moosavi2016deepfool,gong2024adversarial}. In the early stages of research, adversarial attacks had to be customized for each specific sample. However, with the evolution of related studies, universal perturbation~\cite{Moosavi-Dezfooli_2017_CVPR} attacks were introduced, aiming to find perturbations effective across multiple samples rather than tailored to individual instances. Research on universal perturbation attacks seeks to expose vulnerabilities in models, prompting designers to enhance their robustness to withstand a broader range of adversarial challenges.

\textbf{Adversarial Attacks in ReID.}
Some ReID attack methods have been proposed, with current research predominantly focusing on RGB-RGB matching. These methods mainly include: Metric-FGSM~\cite{bai2020adversarial} extends some techniques, inspired by classification attacks, into a category known as metric attacks. These encompass Fast Gradient Sign Method (FGSM)~\cite{fgsm}, Iterative FGSM (IFGSM), and Momentum IFGSM (MIFGSM)~\cite{dong2018boosting}. The Furthest-Negative Attack (FNA)~\cite{bouniot2020vulnerability} integrates hard sample mining~\cite{hermans2017defense} and triple loss to employ pushing and pulling guides. These guides guide image features towards the least similar cluster while moving away from other similar features. Deep Mis-Ranking (DMR)~\cite{wang2020transferable} utilizes a multi-stage network architecture to pyramidally extract features at different levels, aiming to derive general and transferable features for adversarial perturbations. Gong et al.~\cite{gong2022person} proposed a local transformation attack (LTA) method specifically aimed at attacking color features without requiring additional reference images, and discussed effective defense strategies against current ReID attacks. The Opposite-direction Feature Attack (ODFA)~\cite{zheng2023u} exploits feature-level adversarial gradients to generate examples that guide features in the opposite direction with an artificial guide. Yang et al.~\cite{yang2023towards} introduced a combined attack named Col.+Del., which integrates UAP-Retrieval~\cite{li2019universal} with color space perturbations~\cite{laidlaw2019functional}. While this method also explores universal perturbations in ReID, its generality is limited due to the inability to leverage color information in cross-modality problems and the lack of a mechanism for associating different modality information. In contrast to the aforementioned approaches, our focus lies on addressing cross-modality challenges.

\begin{algorithm}
	\caption{Procedure of CMPS attack}
	\label{alg:cmps}
	\begin{algorithmic}[1]
		\State \textbf{Input:} Visible images $I_{RGB}$ and infrared (or thermal) images $I_{ir}$ from dataset $S$, cross-modality ReID model $f$ trained on $S$, adversarial bound $\epsilon$, momentum value $\theta$, iteration step size $\alpha$, iteration epoch $\textit{iter\_epoch}$.
		\State \textbf{Output:} Cross-modality universal perturbation $\eta$.
		\State Initialize $\eta$ with random noise $\eta \leftarrow Rand(0,1)$, $\Delta^0 = 0$.
		\For{$i$ in $\textit{iter\_epoch}$}
		\Repeat
		\State Sample a mini-batch of visible images $I_{RGB}$ and infrared (or thermal) images $I_{ir}$ with $n$ samples
		\State $\hat{I}_{RGB} \leftarrow I_{RGB} + \eta$
		\State Use infrared images to compute the triplet loss $L_{RGB}$ for visible images (Eq.~\ref{f5})
		\State Compute gradient $\Delta_{RGB}$ of $L_{RGB}$ w.r.t. $\eta$:
		\State $\Delta_{RGB} \leftarrow \theta \cdot \Delta^{i-1} + \frac{\partial L_{RGB}}{\partial \eta}$
		\State Update perturbation $\eta$:
		\State $\eta \leftarrow \text{clip}(\eta + \alpha \cdot \text{sign}(\Delta_{RGB}), -\epsilon, \epsilon)$
		\State $\hat{I}_{ir} \leftarrow I_{ir} + \eta$
		\State Use visible images to compute the triplet loss $L_{ir}$ for infrared images (Eq.~\ref{f9})
		\State Compute gradient $\Delta_{ir}$ of $L_{ir}$ w.r.t. $\eta$:
		\State $\Delta^i \leftarrow \theta \cdot \Delta_{RGB} + \frac{\partial L_{ir}}{\partial \eta}$
		\State Update perturbation $\eta$:
		\State $\eta \leftarrow \text{clip}(\eta + \alpha \cdot \text{sign}(\Delta^i), -\epsilon, \epsilon)$
		\Until{all mini-batches are processed}
		\EndFor
		\State \Return $\eta$
	\end{algorithmic}
	\label{algorithm}
\end{algorithm}

\section{Methodology}
In this section, we introduce a universal perturbation designed for cross-modality attacks, referred to as the Cross-Modality Perturbation Synergy (CMPS) attack. Considering the significant differences between different modalities, we propose a attack augmentation method to bridge the gap between modalities, aiding in enhancing the perturbation's universality across different modalities. Our objective in addressing this problem is to find a universal adversarial perturbation, denoted as $\eta$, capable of misleading the retrieval ranking results of cross-modality ReID models. The adversarial operation involves adding $\eta$ to a query image $I$. The perturbed query image, denoted as $I_{adv} = I + \eta$, is then used to retrieve from the gallery and deceive the cross-modality ReID model $f$. The algorithm is summarized in Alg.~\ref{algorithm}.

\subsection{Overall Framework}

In Fig.~\ref{pipline}, we illustrate the overall framework of the proposed CMPS attack. During the training phase, we optimize $\eta$ using our cross-modality attack augmentation method, which leverages images from different modalities to bridge their inherent differences and enhance cross-modality universality. In the attack phase, the optimized $\eta$ deceives reID models, leading to inaccurate ranking lists. Section 3.2 outlines the framework and overall optimization objective, providing a macro-level overview. Section 3.4 delves into the specific process of perturbation optimization across different modalities.

\subsection{Optimizing Loss Functions for Attacking}
Our study aims to deceive cross-modality ReID models using a universal perturbation. We have specifically designed a triplet loss tailored for our proposed attack method, which can correlate different modalities and influence the distance relationships between images from different modalities. 

We follow the approach of~\cite{li2019universal} to optimize the perturbation using cluster centroids. This method directly impacts the similarity between pedestrian identities in the ReID model's feature space (rather than the similarity between individual samples), making it more effective. Subsequently, leveraging the acquired cluster centroids, we apply our triplet loss to distort the pairwise relations between pedestrian identities. This process can be represented as follows:

\begin{equation}
	\resizebox{0.6\linewidth}{!}{
		$\begin{aligned}
			L &= \max \left[ \left( \lVert C^n_{g} - f^{adv}_{RGB} \rVert_2 - \lVert C^p_{ir} - f^{adv}_{RGB} \rVert_2 + \rho \right), 0 \right] \\
			&+ \max \left[ \left( \lVert C^n_{ir} - f^{adv}_{g} \rVert_2 - \lVert C^p_{RGB} - f^{adv}_{g} \rVert_2 + \rho \right), 0 \right] \\
			&+ \max \left[ \left( \lVert C^n_{RGB} - f^{adv}_{ir} \rVert_2 - \lVert C^p_{g} - f^{adv}_{ir} \rVert_2 + \rho \right), 0 \right].
		\end{aligned}$
	}
	\label{loss}
\end{equation}

\begin{figure}[t]
	\resizebox{\linewidth}{!}{
		\begin{minipage}[b]{0.46\linewidth}
			\centering
			\vspace{-2pt} 
			\includegraphics[height=5cm]{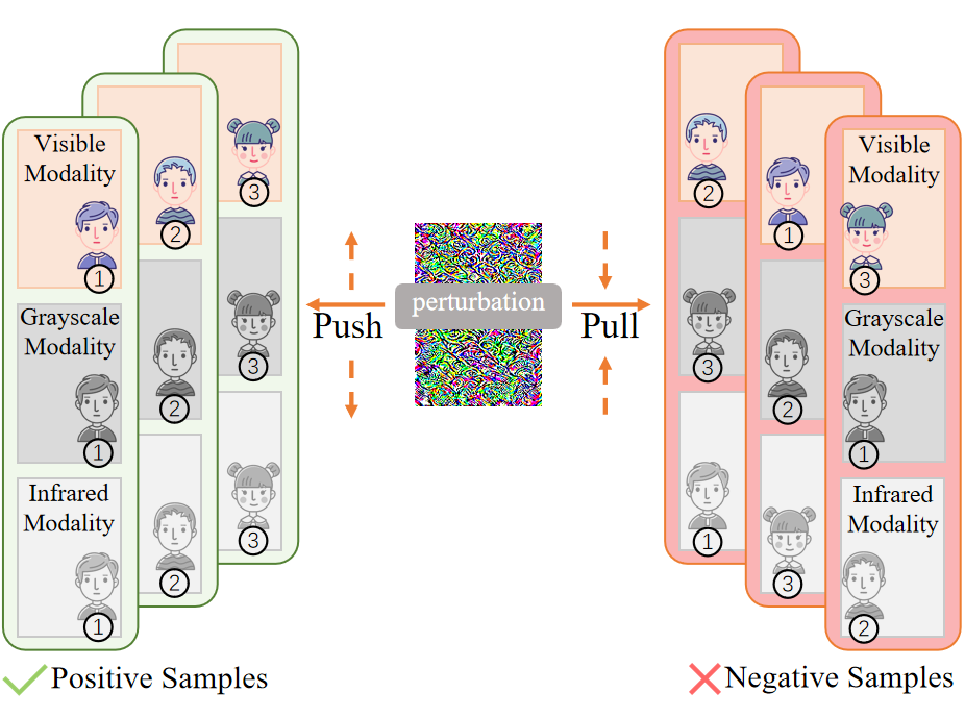} 
			\caption{Schematic illustration of triplet relationship-guided universal perturbation learning for cross-modality ReID.}
			\label{tri}
		\end{minipage}
		\hspace{6pt} 
		\begin{minipage}[b]{0.48\linewidth}
			\centering
			\vspace{-2pt} 
			\includegraphics[height=5cm]{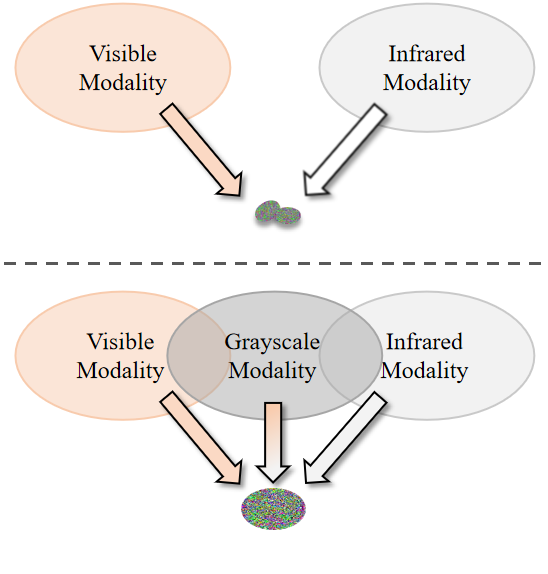} 
			\caption{Cross-modality attack augmentation: bridging gap between visible and non-visible (infrared) modalities with grayscale.}
			\label{aug-method}
		\end{minipage}
	}
\end{figure}

As shown in  Fig.~\ref{tri}, the loss function mentioned above fully leverages the triplet-wise relationships across different modality. Through this loss, we are able to pull the negative samples of each modality closer to the adversarial samples and push the positive samples of each modality away from the adversarial samples. Here, $C^p_{RGB}$ and $C^n_{RGB}$ represent the cluster centroids of the positive samples to push and negative samples to pull, respectively, in the original visible (RGB) image feature space of the training data. Similar definitions apply to other modalities. $f^{adv}_{RGB}$, $f^{adv}_{g}$, and $f^{adv}_{ir}$ denote the perturbed features of the disturbed image in the visible, grayscale, and infrared (or thermal) modalities, respectively.

\subsection{Cross-Modality Attack Augmentation Method}

Intuitively, as illustrated in Fig.~\ref{aug-method}, maximizing the overlap of common factors across different modalities facilitates the capture of shared features by the learned perturbation. Grayscale images, being inherently homogeneous, serve as effective mediators between diverse modalities. Consequently, we introduce random grayscale transformations into adversarial attack methods, referred to as Cross-Modality Attack Augmentation. This approach guides cross-modality perturbations by leveraging homogeneous grayscale images sourced from diverse modalities. The primary objective is to explore the underlying structural relationships across heterogeneous modalities.

The process of grayscale transformation can be represented as follows:

\begin{equation}
	t(R,G,B) = 0.299R+0.587G+0.114B,
\end{equation}
The function t(·) represents the grayscale transformation using ITU-R BT.601-7 standard weights, combining the RGB channels of each pixel into a single grayscale channel. From this, we construct a 3-channel grayscale image $x_g$ by replicating the grayscale channel:
\begin{equation}
	x_g = [	t(R,G,B),t(R,G,B),t(R,G,B)].
\end{equation}

\subsection{Cross-Modality Perturbation Synergy Attack}
To synergistically utilize gradient information from diverse modalities for perturbation optimization, narrow the gap between different modalities to better capture shared knowledge, we adopt the following training process to generate a universal perturbation:


\textbf{(1) Learning the visible modality.} For a given batch of visible images with \( n \) samples, we extract and perturb their features using the cross-modality ReID model. We update the temporary perturbation \( \eta \) iteratively using Momentum-Inertia Stochastic Gradient Descent (MI-SGD), expressed as:

\begin{equation}
	\begin{aligned}
		L_{RGB}(f^{adv}_{RGB}, \eta) &= \max \left[ \left( \lVert C^n_{g} - f^{adv}_{RGB} \rVert_2 \right. \right. \\
		&\quad \left. - \left. \lVert C^p_{ir} - f^{adv}_{RGB} \rVert_2 + \rho \right), 0 \right],
	\end{aligned}
	\label{f5}
\end{equation}

\begin{equation}
	\begin{aligned}
		\Delta_{RGB} = \theta \Delta'_{ir} + \frac{\nabla_{\eta}L_{RGB}}{\| \nabla_{\eta}L_{RGB} \|_1 },
	\end{aligned}
	\label{loss1}
\end{equation}

\begin{equation}
	\begin{aligned}
		\eta = \text{clip}(\eta + \alpha \cdot \text{sign}(\Delta_{RGB}), -\varepsilon,\varepsilon).
	\end{aligned}
	\label{loss1}
\end{equation}

Here, \( \theta \) represents the momentum value (set as \( \theta = 1 \)), and $\Delta'_{ir}$ is derived from the previous iteration. The iteration step size is denoted by $\alpha$ (set as $\alpha = \frac{\epsilon}{12}$ ), where \( \epsilon \) is the adversarial bound ($ \epsilon = 8 $, unless otherwise specified). We set the margin $\rho=0.5$ in our triplet loss.

\textbf{(2) Learning the grayscale modality.}
This part is executed through data augmentation. It is not considered as a separate module and is therefore not explicitly listed in Alg.~\ref{algorithm}. Specifically, during the perturbation learning process, we randomly transform visible or infrared (or thermal) images into homogeneous grayscale images, participating in the iterative optimization of adversarial perturbations. It is employed to bridge the gap between different modalities, thereby improving the universality of the perturbation across diverse modalities. In order to investigate the impact of different grayscale conversion probabilities on attack performance, we conducted a series of ablation experiments. For details, please refer to Fig.~\ref{aug2} in supplementary material.

\textbf{(3) Learning the infrared (or thermal) modality.}
This step is similar to (1). We utilize the infrared (or thermal) images to learn the perturbation $\eta$ with the our loss functions:

\begin{equation}
	\begin{aligned}
		L_{ir}(f^{adv}_{ir}, \eta) &= \max \left[ \left( \lVert C^n_{RGB} - f^{adv}_{ir} \rVert_2 \right. \right. \\
		&\quad \left. - \left. \lVert C^p_{g} - f^{adv}_{ir} \rVert_2 + \rho \right), 0 \right],
	\end{aligned}
\label{f9}
\end{equation}

\begin{equation}
	\begin{aligned}
		\Delta_{ir} = \theta \Delta_{RGB} + \frac{\nabla_{\eta}L_{ir}}{\| \nabla_{\eta}L_{ir} \|_1 },
	\end{aligned}
	\label{loss1}
\end{equation}
\begin{equation}
	\begin{aligned}
		\eta = \text{clip}(\eta +	 \alpha \cdot \text{sign}(\Delta_{ir}), -\varepsilon,\varepsilon).
	\end{aligned}
	\label{loss1}
\end{equation}
Here, $\Delta_{RGB}$ derived from step (1). The main difference compared to the previous step lies in the perturbation applied to the input and the gradients related to momentum. 

\textbf{Theoretical Analysis.} In traditional optimization, optimizing for one modality can render the perturbation suboptimal for the other, leading to a bias toward a single modality. In contrast, the proposed aggregated optimization method jointly optimizes both modalities, ultimately identifying a universal perturbation that enhances cross-modality attack performance. In the supplementary material~\ref{proof}, we provide a mathematical analysis demonstrating the effectiveness of this method compared to traditional attack methods that lack intrinsic correlations between different modalities.

\section{Experiments}
In this section, we compare our approach with several methods, including traditional classification attack methods FGSM~\cite{fgsm} and PGD~\cite{madry2018towards}, traditional metric attack methods like Metric-FGSM~\cite{bai2020adversarial}, as well as state-of-the-art ReID attack methods such as LTA~\cite{gong2022person} \footnote{The LTA code is available at: https://github.com/finger-monkey/LTA\_and\_joint-defence
}, ODFA\cite{zheng2023u} and Col.+Del.\cite{yang2023towards}.

\textbf{Datasets}. We evaluate our proposed method on two commonly used cross-modality ReID datasets: SYSU-MM01~\cite{wu2017rgb}, RegDB~\cite{nguyen2017person} and LLCM~\cite{zhang2023diverse}. SYSU-MM01 is a large-scale dataset with 395 training identities, captured by 6 cameras (4 RGB, 2 near-infrared) on the SYSU campus. It comprises 22,258 visible and 11,909 near-infrared images. The testing set consists of 95 identities with two evaluation settings. The query sets include 3803 images from two IR cameras. We conduct ten trials following established methods~\cite{wang2019learning} and report the average retrieval performance. Please refer to~\cite{wu2017rgb} for the evaluation protocol. RegDB~\cite{nguyen2017person} is a smaller-scale dataset with 412 identities, each having ten visible and ten thermal images. we randomly select 206 identities (2,060 images) for training and use the remaining 206 identities (2,060 images) for testing. LLCM is a dataset designed specifically for cross-modality ReID in low-light environments. Compared to other datasets, its diverse scenarios and low-light conditions present greater challenges for attackers. This complexity and uncertainty make adversarial attacks more difficult to execute. We assess our model in two retrieval scenarios: visible-thermal and thermal-visible performance. 

\textbf{Evaluation Metrics}. Following existing works~\cite{zheng2015scalable}, we employ Rank-k precision and Cumulative Matching Characteristics (CMC) and mean Average Precision (mAP) as evaluation metrics. Rank-1 represents the average accuracy of the top-ranked result corresponding to each cross-modality query image. mAP represents the mean average accuracy, where the query results are sorted based on similarity, and the closer the correct result is to the top of the list, the higher the precision. Please note that, for adversarial attacks, a lower accuracy indicates a more successful attack.

\begin{table}[h!]
	\centering
	\caption{Results for attacking cross-modality ReID systems on the SYSU~\cite{wu2017rgb} dataset. It reports on visible images querying infrared images and vice versa. Rank at \( r \) accuracy (\%) and mAP (\%) are reported. For the "Visible to Infrared" scenario, we used the all-search mode. For the "Infrared to Visible" scenario, we used the indoor-search mode. }
	\label{tab:table1}
	\resizebox{\textwidth}{!}{ 
		\begin{tabular}{llccccccccc}
			\toprule
			\multicolumn{2}{c}{Settings} & \multicolumn{4}{c}{Visible to Infrared} & \multicolumn{4}{c}{Infrared to Visible} \\
			\cmidrule(r){1-2} \cmidrule(lr){3-6} \cmidrule(l){7-10}
			Method & Venue & \( r = 1 \) & \( r = 10 \) & \( r = 20 \) & mAP & \( r = 1 \) & \( r = 10 \) & \( r = 20 \) & mAP \\
			\midrule
			AGW baseline~\cite{ye2021deep} & TPAMI 2022 & 47.50 & 84.39 & 92.14 & 47.65 & 54.17 & 91.14 & 95.98 & 62.97 \\
			FGSM attack~\cite{fgsm} & ICLR 2015 & 42.64 & 81.21 & 89.32 & 43.67 & 48.05 & 86.73 & 92.11 & 53.22 \\
			PGD attack~\cite{madry2018towards} & ICLR 2018 & 39.14 & 76.80 & 85.42 & 40.91 & 43.68 & 82.54 & 89.14 & 48.56 \\
			M-FGSM attack~\cite{bai2020adversarial} & TPAMI 2020 & 25.79 & 49.04 & 57.96 & 19.24 & 20.56 & 38.91 & 46.35 & 15.89 \\
			LTA attack~\cite{gong2022person} & CVPR 2022 & 8.42 & 21.25 & 27.98 & 9.16 & 20.92 & 32.18 & 36.80 & 15.24 \\
			ODFA attack~\cite{zheng2023u} & IJCV 2023 & 25.43 & 47.49 & 56.38 & 19.00 & 14.62 & 29.92 & 36.42 & 11.35 \\
			Col.+Del. attack~\cite{yang2023towards} & TPAMI 2023 & 3.23 & 14.48 & 20.15 & 3.27 & 4.12 & 16.85 & 21.27 & 3.89 \\
			Our attack & NeurIPS 2024 & 1.11 & 8.67 & 16.14 & 1.41 & 1.31 & 7.47 & 10.36 & 1.23 \\
			\midrule 
			DDAG baseline~\cite{eccv20ddag} & ECCV 2020 & 54.75 & 90.39 & 95.81 & 53.02 & 61.02 & 94.06 & 98.41 & 67.98 \\
			FGSM attack~\cite{fgsm} & ICLR 2015 & 48.27 & 86.02 & 91.34 & 49.55 & 53.87 & 90.15 & 94.58 & 57.84 \\
			PGD attack~\cite{madry2018towards} & ICLR 2018 & 50.62 & 88.30 & 93.12 & 51.89 & 56.10 & 91.54 & 96.13 & 59.22 \\
			M-FGSM attack~\cite{bai2020adversarial} & TPAMI 2020 & 28.36 & 52.47 & 60.76 & 23.11 & 24.85 & 40.74 & 49.23 & 18.40 \\
			LTA attack~\cite{gong2022person} & CVPR 2022 & 10.54 & 23.08 & 30.47 & 12.28 & 18.93 & 34.12 & 41.52 & 15.04 \\
			ODFA attack~\cite{zheng2023u} & IJCV 2023 & 27.75 & 50.26 & 59.14 & 22.30 & 17.62 & 32.64 & 40.03 & 14.83 \\
			Col.+Del. attack~\cite{yang2023towards} & TPAMI 2023 & 4.28 & 16.12 & 21.36 & 3.97 & 6.28 & 19.53 & 25.61 & 5.21 \\
			Our attack &NeurIPS 2024 & 1.62 & 7.59 & 14.46 & 1.84 & 1.45 & 7.71 & 10.72 & 1.25 \\
			\bottomrule
		\end{tabular}
	}
	\label{sysu}
\end{table}

\begin{table}[h!]
	\centering
	\caption{Results for attacking cross-modality ReID systems on the RegDB~\cite{nguyen2017person} dataset. It reports on visible images querying thermal images and vice versa. Rank at \( r \) accuracy (\%) and mAP (\%) are reported.}
	\label{tab:table1}
	\resizebox{\textwidth}{!}{ 
		\begin{tabular}{llccccccccc}
			\toprule
			\multicolumn{2}{c}{Settings} & \multicolumn{4}{c}{Visible to Thermal} & \multicolumn{4}{c}{Thermal to Visible} \\
			\cmidrule(r){1-2} \cmidrule(lr){3-6} \cmidrule(l){7-10}
			Method & Venue & \( r = 1 \) & \( r = 10 \) & \( r = 20 \) & mAP & \( r = 1 \) & \( r = 10 \) & \( r = 20 \) & mAP \\
			\midrule
			AGW baseline~\cite{ye2021deep}  & TPAMI 2022 & 70.05 & 86.21 & 91.55 & 66.37 & 70.49 & 87.21 & 91.84 & 65.90 \\
			FGSM attack~\cite{fgsm} & ICLR 2015 & 66.79 & 83.14 & 88.46 & 61.05 & 65.42 & 81.98 & 87.20 & 60.12 \\
			PGD attack~\cite{madry2018towards} & ICLR 2018 & 62.14 & 80.28 & 85.10 & 57.34 & 63.71 & 78.82 & 84.05 & 58.42 \\	
			M-FGSM attack~\cite{bai2020adversarial} & TPAMI 2020 & 29.34 & 52.90 & 61.44 & 23.35 & 23.64 & 40.36 & 48.61 & 18.57 \\
			LTA attack~\cite{gong2022person} & CVPR 2022 & 12.65 & 25.24 & 34.02 & 12.80 & 10.51 & 22.93 & 31.79 & 9.74 \\
			ODFA attack~\cite{zheng2023u} & IJCV 2023 & 28.57 & 51.42 & 60.58 & 21.84 & 17.26 & 33.27 & 42.92 & 15.27 \\
			Col.+Del. attack~\cite{yang2023towards} & TPAMI 2023 & 5.12 & 16.83 & 22.10 & 4.94 & 4.92 & 14.47 & 23.04 & 4.86 \\
			Our attack & NeurIPS 2024 & 2.29 & 9.06 & 18.35 & 3.92 & 1.93 & 11.44 & 19.30 & 3.46 \\
			\midrule 
			DDAG baseline~\cite{eccv20ddag} & ECCV 2020 & 69.34 & 86.19 & 91.49 & 63.46 & 68.06 & 85.15 & 90.31 & 61.80 \\
			FGSM attack~\cite{fgsm} & ICLR 2015 & 61.83 & 80.12 & 86.47 & 55.78 & 60.94 & 78.35 & 84.09 & 56.91 \\
			PGD attack~\cite{madry2018towards} & ICLR 2018 & 64.58 & 81.39 & 87.20 & 58.45 & 62.17 & 79.02 & 85.27 & 57.69 \\	
			M-FGSM attack~\cite{bai2020adversarial} & TPAMI 2020 & 30.86 & 54.16 & 61.98 & 24.01 & 25.83 & 42.12 & 49.76 & 19.33 \\
			LTA attack~\cite{gong2022person} & CVPR 2022 & 11.65 & 23.20 & 32.73 & 11.41 & 9.76 & 21.53 & 29.96 & 9.23 \\
			ODFA attack~\cite{zheng2023u} & IJCV 2023 & 29.64 & 52.74 & 60.74 & 23.88 & 24.06 & 39.75 & 46.25 & 18.64 \\
			Col.+Del. attack~\cite{yang2023towards} & TPAMI 2023 & 4.68 & 13.55 & 18.57 & 4.39 & 4.23 & 12.75 & 20.82 & 4.05 \\
			Our attack & NeurIPS 2024 & 1.33 & 10.28 & 19.06 & 3.79 & 1.35 & 9.52 & 17.52 & 3.19 \\
			\bottomrule
		\end{tabular}
	}
	\label{regdb}
\end{table}

\subsection{Performance on Cross-Modality ReID}
We used AGW~\cite{ye2021deep} and DDAG~\cite{eccv20ddag} as baseline models for testing on the RegDB and SYSU cross-modality ReID datasets. AGW (Attention Generalized mean pooling with Weighted triplet loss) enhances the learning capability of crucial features by integrating non-local attention blocks, learnable GeM pooling, and weighted regularization triplet loss. DDAG (Dynamic Dual-Attentive Aggregation) improves feature learning by combining intra-modality weighted-part attention and cross-modality graph structured attention, considering both part-level and cross-modal contextual cues. Additionally, we use DEEN~\cite{zhang2023diverse} (Diverse Embedding Expansion Network) as baseline models for testing on the LLCM~\cite{zhang2023diverse} cross-modality ReID datasets. The core idea of DEEN is to enhance the feature representation capability by introducing a diversity embedding mechanism. The network expands the embedding space, allowing features from visible and infrared images to align better in a high-dimensional space, thereby improving the accuracy of cross-modality matching.

The experiments encompass two scenarios: 1) Perturbing visible images (query) to disrupt the retrieval of infrared or thermal non-visible images (gallery). This is denoted as "Visible to Infrared" in Tab.\ref{sysu} and "Visible to Thermal" in Tab.\ref{regdb}. 2) Perturbing infrared or thermal non-visible images (query) to interfere with the retrieval of visible images (gallery). This is indicated as "Infrared to Visible" in Tab.\ref{sysu} and "Thermal to Visible" in Tab.\ref{regdb}.

From Tab.\ref{sysu}, it can be seen that the proposed method reduces the rank-1 accuracy to below 2\% in both the 'Visible to Infrared' and 'Infrared to Visible' cases. Similarly, from Tab.\ref{regdb}, the rank-1 accuracy drops below 3\% in both the 'Visible to Thermal' and 'Thermal to Visible' scenarios. In contrast, traditional metric-based attacks, such as Metric-FGSM (M-FGSM)\cite{bai2020adversarial}, LTA~\cite{gong2022person} and ODFA\cite{zheng2023u}, lead to attacked models with significantly higher rank-1 accuracy, whereas traditional classification attacks (such as FGSM~\cite{fgsm} and PGD~\cite{madry2018towards}) perform even worse, with rank-1 accuracy remaining over 60\%. This is because ReID relies on metric learning for feature matching rather than category classification, requiring attacks specifically tailored for metric learning.
These results indicate that, compared to traditional methods that optimize perturbations separately for each modality without considering the inherent correlations between different modalities, our proposed approach demonstrates significant attacking effectiveness across different modalities.

\begin{table}[h!]
	\centering
	\caption{Results for attacking cross-modality ReID systems on the LLCM~\cite{zhang2023diverse} dataset. It reports on visible images querying thermal images and vice versa. Rank at \( r \) accuracy (\%) and mAP (\%) are reported.}
	\label{tab:table1}
	\resizebox{\textwidth}{!}{ 
		\begin{tabular}{llccccccccc}
			\toprule
			\multicolumn{2}{c}{Settings} & \multicolumn{4}{c}{Visible to Infrared} & \multicolumn{4}{c}{Infrared to Visible} \\
			\cmidrule(r){1-2} \cmidrule(lr){3-6} \cmidrule(l){7-10}
			Method & Venue & \( r = 1 \) & \( r = 10 \) & \( r = 20 \) & mAP & \( r = 1 \) & \( r = 10 \) & \( r = 20 \) & mAP \\
			\midrule
			DEEN baseline~\cite{ye2021deep}  & CVPR 2023 &62.53  &90.31  &94.73  &65.84  &54.96  &84.92  &90.91  &62.95  \\
			M-FGSM attack~\cite{bai2020adversarial} & TPAMI 2020 &28.48  &64.92  &75.12  &32.88  &25.64  &61.45  &78.31  &30.46  \\
			LTA attack~\cite{gong2022person} & CVPR 2022  &15.16  &56.42  &67.53  &21.47  &19.54  &58.25  &70.72  &24.86  \\
			ODFA attack~\cite{zheng2023u} & IJCV 2023  &26.34  &65.24  &76.92  &30.85  &23.73  &62.46  &73.57  &29.63  \\
			Col.+Del. attack~\cite{yang2023towards} & TPAMI 2023  &8.61  &22.73  &36.07  &15.72  &9.13  &20.76  &38.02  &16.31  \\
			Our attack & NeurIPS 2024  &5.83  &18.14  &27.56  &12.47  &6.42  &19.53  &28.54  &12.23  \\
			\bottomrule
		\end{tabular}
	}
	\label{LLCM}
\end{table}

\textbf{Comparison with State-of-the-Art.} Col.+Del., as a universal perturbation method, was fairly compared by first optimizing with one modality's dataset and then fine-tuning with the other modality. Since universal perturbations capture shared patterns across the entire data distribution, Col.+Del. is capable of achieving some level of attack effectiveness in cross-modality scenarios. However, by comparing Tab.\ref{sysu}, Tab.\ref{regdb}, and Tab.\ref{LLCM}, we observe that although Col.+Del. performs better than other methods, its effectiveness is still noticeably limited due to the lack of intrinsic correlation mechanisms between modalities. Moreover, as shown in Fig.\ref{trans}, our method outperforms Col.+Del. in transfer attacks across different baselines in cross-modality ReID. The conclusions from these experiments are as follows: 1) In cross-modality attacks, Col.+Del. demonstrates the feasibility of universal perturbations. However, its performance is limited by its failure to account for modality differences and inherent correlations. 2) Our method better bridges the gap between different modalities, more effectively capturing shared features across them.

\subsection{Transferability of CMPS}
From Fig.\ref{trans} in supplemental material, the results of the proposed method's transfer attacks on two baseline models, AGW and DDAG, can be observed. For example, on the SYSU dataset, the original attack result of the proposed method on DDAG is mAP=1.84\% (refer to Tab.~\ref{sysu}). When the perturbation is transferred from AGW to DDAG, the attack result becomes mAP=3.41\%. This indicates that the proposed attack method exhibits good generalization across different models, and thus, the attack performance does not degrade significantly. This consistent result is observed on both the RegDB and SYSU datasets. Similarly, in Fig.\ref{trans2} of the supplemental material, we evaluate the cross-dataset transferability of perturbations in comparison with Col.+Del. The results demonstrate a significant advantage of our method. Additionally, we conducted adversarial transferability experiments on IDE~\cite{IDE}, PCB~\cite{PCB}, and ResNet18~\cite{resnet}. The rank-1 transfer attack success rates are presented in Tab.\ref{Transf}. It can be observed that our method consistently achieves higher transfer attack success rates across all model combinations compared to Col.+Del., indicating that our method demonstrates stronger robustness in generating more universal adversarial perturbations.

\begin{table}[h]\small
	\centering
	\caption{Comparison of transfer attack success rates between our method and Col.+Del. across models, with higher values indicating better transferability.}
	\begin{tabular}{lccc}
		\toprule
		Source \textbackslash Target Model & IDE (Ours/Col.+Del.) & PCB (Ours/Col.+Del.) & ResNet18 (Ours/Col.+Del.) \\
		\midrule
		IDE~\cite{IDE}       & 98.7\% / 94.3\% & 84.5\% / 81.2\% & 87.4\% / 86.1\% \\
		PCB~\cite{PCB}       & 85.1\% / 80.4\% & 97.6\% / 92.8\% & 88.3\% / 85.7\% \\
		ResNet18~\cite{resnet}  & 81.0\% / 78.5\% & 77.5\% / 74.9\% & 98.2\% / 95.6\% \\
		\bottomrule
	\end{tabular}
\label{Transf}
\end{table}

\subsection{Ablation Study}
Our method is implemented based on UAP-Retrieval~\cite{li2019universal}. To validate the effectiveness of the proposed method, we conducted experiments by adding augmentation (Cross-Modality Attack Augmentation) and CMPS to the baseline. Results with AGW baseline model are reported in Tab.~\ref{ablation}. The No.1 line represents the UAP-Retrieval algorithm. In the table, 'Aug' indicates the use of the Cross-Modality Attack Augmentation proposed in this paper.

\textbf{The effectiveness of CMPS}. 
Comparing No.1 with No.3 and No.4, we observe the following: 1) The direct use of UAP-Retrievals yields limited performance. 2) Training with the CMPS strategy proposed in this paper consistently improves the performance of attack results and the universality of learned perturbations.

\begin{table}[t]
	\centering
	\caption{Ablation studies on the AGW baseline. 'Aug' denotes the cross-modality attack augmentation method proposed in this paper.}
	\begin{tabular}{cccccccc}
		\toprule
		No. & \multicolumn{2}{c}{RegDB} & \multicolumn{2}{c}{SYSU} & \multicolumn{1}{c}{Aug} & \multicolumn{1}{c}{CMPS} \\
		\cmidrule(r){2-3} \cmidrule(r){4-5}
		& mAP & rank-1 & mAP & rank-1 & & \\
		\midrule
		1 & 6.87 & 5.53 & 4.76 & 5.09 & $\times$ & $\times$ \\
		2 & 5.11 & 4.02 & 3.85 & 4.37 & $\checkmark$ & $\times$ \\
		3 & 3.98 & 2.17 & 3.42 & 3.82 & $\times$ & $\checkmark$ \\
		4 & 3.46 & 1.93 & 1.23 & 1.31 & $\checkmark$ & $\checkmark$ \\
		\bottomrule
	\end{tabular}
	\label{ablation}
\end{table}

\textbf{The effectiveness of augmentation method}. 
Our approach includes cross-modality attack augmentation. Comparing results of No.1, No.2, and No.4 shows its benefits. For example, on the RegDB dataset, augmentation (No.2) reduces mAP from 6.87\% to 5.11\%, 1.76\% lower than without augmentation (No.1). Similarly, with CMPS, mAP drops from 3.98\% to 3.46\% (No.4), a 0.52\% decrease compared to No.3. These findings suggest that using appropriate augmentation enhances cross-modality ReID adversarial attacks' universality. If not specified, our experiments default to using CMPS augmentation. Fig.~\ref{aug2} in the supplementary materials displays the experimental results of our augmentation performed at different probabilities. It can be observed that when the probability value is around 20\%, it achieves optimal effectiveness in assisting the attack. If not specified, a probability value of 20\% for augmentation is used by default in experiments.

\textbf{Impact of adversarial boundary size.}
We conducted an ablation study on different adversarial boundary sizes ($\epsilon$), as shown in the supplementary material ~\ref{eps}. In practical applications, $\epsilon$ is typically kept moderate to balance perturbation visibility and attack effectiveness. To maintain consistency with previous work ~\cite{yang2023towards}, we set $\epsilon = 8$ for comparison unless otherwise specified.

\section{Conclusion}
In this study, we have proposed a cross-modality attack method known as Cross-Modality Perturbation Synergy (CMPS) attack, aimed at evaluating the security of cross-modality ReID systems. The core idea behind the CMPS attack is to capture shared knowledge between visible and non-visible images to optimize perturbations. Additionally, we proposed a Cross-Modality Attack Augmentation method, utilizing grayscale images to bridge the gap between different modalities, further enhancing the attack performance. Through experiments conducted on the RegDB, SYSU and LLCM datasets, we demonstrated the effectiveness of the proposed method while also revealing the limitations of traditional attack approaches. The primary objective of this study has been to assess the security of cross-modality ReID systems. In future research, on the one hand, we will continue to improve the transferability of cross-modality attacks across different datasets and models; on the other hand, we plan to develop robust ReID methods specifically tailored for cross-modality attacks, aimed at defending against adversarial samples. This study not only contributes to advancing the understanding of the security of cross-modality ReID systems but also provides strong motivation for ensuring the reliability and security of these systems in real-world applications.

\section*{Acknowledgment}

This work was supported in part by the National Natural Science Foundation of China under Grant No.62276222 and the Public Technology Service Platform Project of Xiamen City, Grant No.3502Z20231043.

\newpage
\bibliographystyle{unsrt}
\bibliography{CMPS_neurips_2024_1}

\newcommand{\suptitle}[1]{%
	\begin{center}
		\Large\bfseries #1
	\end{center}
}

\clearpage

\suptitle{Supplemental Material}

\begin{minipage}{\linewidth}
	\tableofcontents  
\end{minipage}
\newpage

\section{Supplemental Experiments}
Our experiments were conducted using three RTX 2080 Ti GPUs, each with 11GB of memory. 

\begin{figure}[h]
	\setlength{\abovecaptionskip}{0.1cm}
	\setlength{\belowcaptionskip}{-0.2 cm}   
	\centering
	\includegraphics[width=0.95\linewidth]{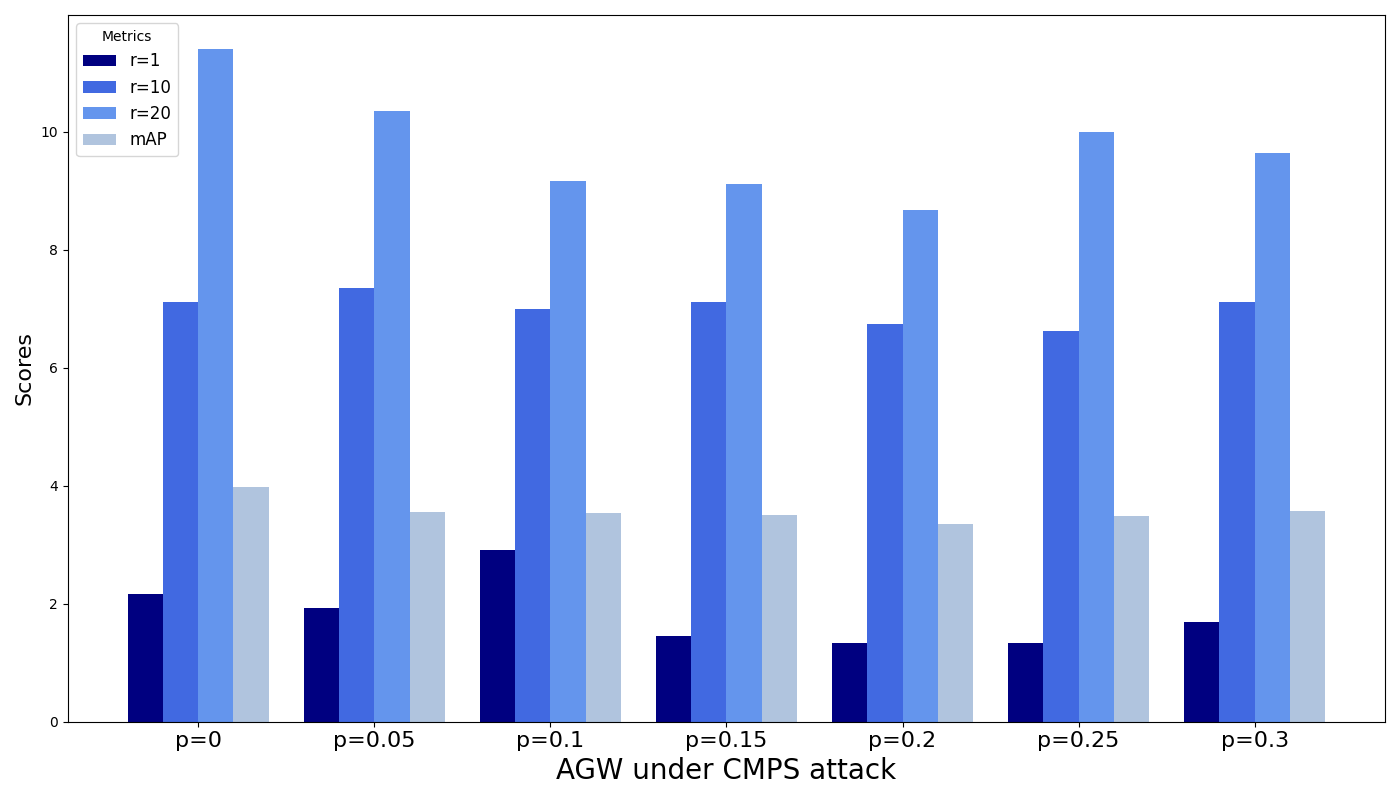}
	\caption{The impact of different grayscale transformation probabilities on attack performance. Lower evaluation metrics indicate higher attack success rates. The experimental results are derived from experiments on the RegDB dataset using AGW as the baseline model for testing.}
	\label{aug2}
\end{figure}

\begin{figure}[h]
	\setlength{\abovecaptionskip}{0.1cm}
	\setlength{\belowcaptionskip}{-0.2 cm}   
	\centering
	\includegraphics[width=1\linewidth, height=0.5\linewidth]{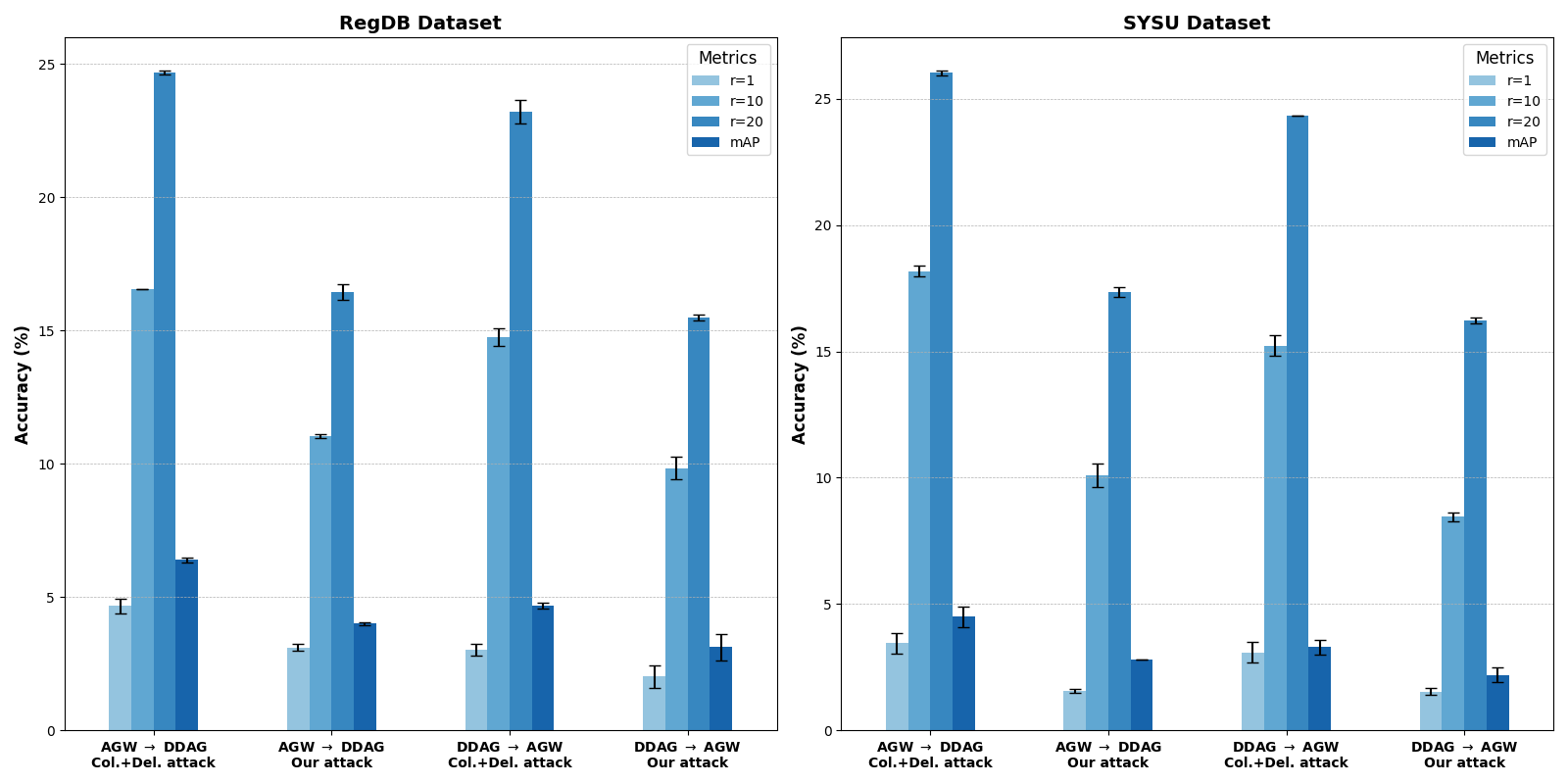}
	\caption{Transferability experiments of the proposed method across different models on the RegDB dataset (visible to thermal). Transferability experiments of the proposed method across different models on the SYSU dataset (visible to Infrared).}
	\label{trans}
\end{figure}

\begin{figure}[h]
	\setlength{\abovecaptionskip}{0.1cm}
	\setlength{\belowcaptionskip}{-0.2 cm}   
	\centering
	\includegraphics[width=1\linewidth]{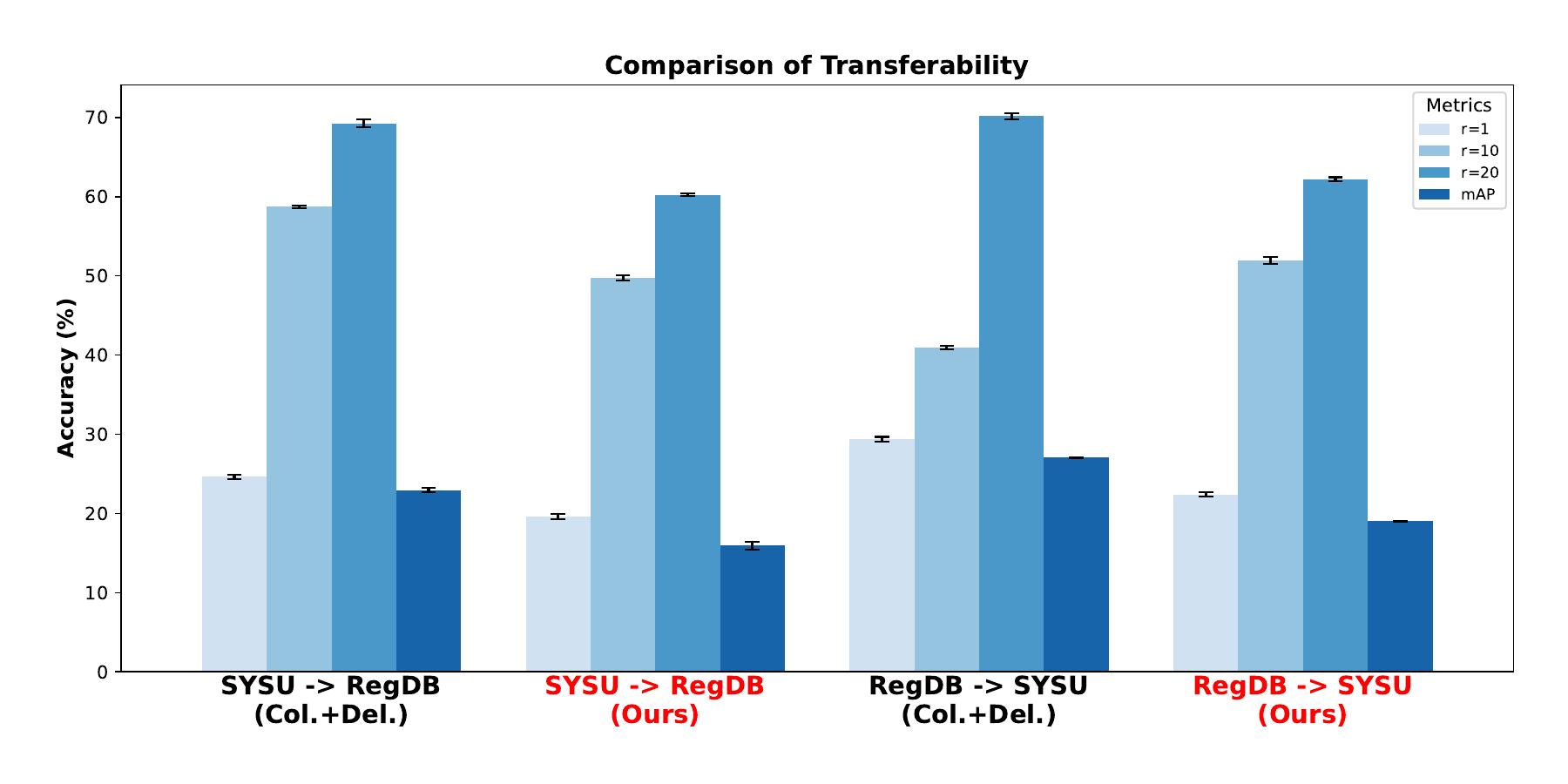}
	\caption{Comparison of Transferability Between Different Methods on Two Cross-Modal Datasets SYSU and RegDB.}
	\label{trans2}
\end{figure}

\begin{table}[h!]
	\centering
	\caption{Using the AGW baseline on the RegDB dataset, we conduct an ablation study to evaluate the impact of the adversarial boundary $\epsilon$ on the effectiveness of the proposed CMPS attack (rank-1 accuracy).}
	\label{tab:adversarial_boundary}
	\begin{tabular}{ccc}
		\toprule
		\textbf{Adversarial Boundary ($\epsilon$)} & \textbf{Visible to Thermal} & \textbf{Thermal to Visible} \\
		\midrule
		\multicolumn{1}{c|}{-}    & \multicolumn{1}{c|}{70.0\%}  & 70.5\%  \\
		\multicolumn{1}{c|}{2}    & \multicolumn{1}{c|}{32.7\%}  & 40.5\%  \\
		\multicolumn{1}{c|}{4}    & \multicolumn{1}{c|}{9.6\%}   & 13.8\%  \\
		\multicolumn{1}{c|}{8}    & \multicolumn{1}{c|}{2.3\%}   & 2.0\%   \\
		\multicolumn{1}{c|}{16}   & \multicolumn{1}{c|}{0.3\%}   & 0.5\%   \\
		\bottomrule
	\end{tabular}
\label{eps}
\end{table}

\clearpage
	
\section{Proof of Method Superioritys}\label{proof}

We design a cross-modality triplet loss to simultaneously optimize two modalities, which effectively captures common features between different modalities and enhances the cross-modality adaptability of universal perturbations.

\subsection{Definition of Cross-Modality Triplet Loss}

The cross-modality triplet loss aims to optimize the model by adjusting the distance relationships among triplet samples (anchor, positive, negative) so that samples of the same identity are closer, while samples of different identities are farther apart. Specifically, given samples $(x_{\text{A}}, x_{\text{P}}, x_{\text{N}})$, where:
\begin{itemize}
	\item $x_{\text{A}}$ is the anchor sample,
	\item $x_{\text{P}}$ is the positive sample with the same identity as the anchor (from a different modality),
	\item $x_{\text{N}}$ is the negative sample with a different identity from the anchor.
\end{itemize}

The triplet loss function is defined as:
\begin{equation}
L_{\text{triplet}} = \max \left(0, D(f(x_{\text{A}}), f(x_{\text{P}})) - D(f(x_{\text{A}}), f(x_{\text{N}})) + \alpha \right)
\end{equation}
where $D(\cdot, \cdot)$ denotes the distance metric (e.g., Euclidean distance), and $\alpha$ is a margin hyperparameter.

Mathematically, given the cross-modality triplet loss:
\begin{equation}
L_{\text{triplet}} = \max \left( \left( \lVert C^n_g - f^{adv}_{RGB} \rVert_2 - \lVert C^p_{ir} - f^{adv}_{RGB} \rVert_2 + \rho \right), 0 \right)
\end{equation}

We can view it as part of the sum of the loss functions for two modalities:
\begin{equation}
\mathcal{L}_A(\eta) = \lVert C^n_g - f^{adv}_{RGB} \rVert_2
\end{equation}
\begin{equation}
\mathcal{L}_B(\eta) = \lVert C^p_{ir} - f^{adv}_{RGB} \rVert_2
\end{equation}

Thus, the overall optimization objective can be expressed as:
\begin{equation}
\eta_{\text{agg}}^* = \arg \min_{\eta} \left( \mathcal{L}_A(\eta) + \mathcal{L}_B(\eta) + \rho \right)
\end{equation}

This form effectively aggregates the losses of different modalities, thereby optimizing the loss functions of different modalities simultaneously, achieving joint optimization of cross-modality data. This approach trains universal perturbations with better generalization capabilities than methods that consider only single-modality information.

\subsection{Proof of Aggregated Optimization Superiority}

Assume we have data from two modalities: modality A and modality B. Let $\mathcal{L}_A(\eta)$ and $\mathcal{L}_B(\eta)$ be the loss functions on modality A and modality B, respectively. The objective of single-modality training is:
\begin{equation}
\min_{\eta} \mathcal{L}_A(\eta) + \mathcal{L}_B(\eta)
\end{equation}

The stepwise optimization method first optimizes $\mathcal{L}_A(\eta)$ and then optimizes $\mathcal{L}_B(\eta)$:
\begin{equation}
\eta^* = \arg \min_{\eta} \mathcal{L}_A(\eta) \rightarrow \eta^{**} = \arg \min_{\eta} \mathcal{L}_B(\eta^*)
\end{equation}

The aggregated optimization of the two loss functions is:
\begin{equation}
\eta_{\text{agg}}^* = \arg \min_{\eta} \left( \mathcal{L}_A(\eta) + \mathcal{L}_B(\eta) \right)
\end{equation}

Using the gradient aggregation method, it can be expressed as:
\begin{equation}
\nabla_{\eta} \mathcal{L}_{\text{agg}} = \nabla_{\eta} \left( \mathcal{L}_A(\eta) + \mathcal{L}_B(\eta) \right)
\end{equation}

Next, we consider the different optimization paths of the two methods.

\subsubsection{Stepwise Optimization Method}

The stepwise optimization method first optimizes the loss function of modality A and then the loss function of modality B. Assume the update rule at iteration $k$ is:
\begin{equation}
\eta^{(k+1)} = \eta^{(k)} - \alpha \nabla_{\eta} \mathcal{L}_A(\eta^{(k)})
\end{equation}

After optimizing the loss function of modality A, the loss function of modality B is optimized:
\begin{equation}
\eta^{(k+1)} = \eta^{(k)} - \alpha \nabla_{\eta} \mathcal{L}_B(\eta^{(k)})
\end{equation}

Since the two optimization processes are separate, this may result in $\eta$ being optimal for modality A but not necessarily for modality B.

\subsubsection{Aggregated Optimization Method}

The aggregated optimization method considers the losses of both modalities in each iteration. Assume the update rule at iteration $k$ is:
\begin{equation}
\eta^{(k+1)} = \eta^{(k)} - \alpha \left( \nabla_{\eta} \mathcal{L}_A(\eta^{(k)}) + \nabla_{\eta} \mathcal{L}_B(\eta^{(k)}) \right)
\end{equation}

In this way, each update considers the losses of both modalities, ensuring that $\eta$ approaches the optimal solution for both modalities.

To further prove that the aggregated optimization method can find a better perturbation $\eta$, we can analyze the existence and uniqueness of the optimal solution.

Assume $\mathcal{L}_A(\eta)$ and $\mathcal{L}_B(\eta)$ are continuously differentiable and convex loss functions. According to convex optimization theory, the optimal solutions of the loss functions exist and are unique.

The optimal solution of the stepwise optimization method is:
\begin{equation}
\eta_{\text{step}}^* = \arg \min_{\eta} \left( \mathcal{L}_A(\eta) + \mathcal{L}_B(\eta^*) \right)
\end{equation}
where $\eta^*$ is the optimal solution of $\mathcal{L}_A(\eta)$.

The optimal solution of the aggregated optimization method is:
\begin{equation}
\eta_{\text{agg}}^* = \arg \min_{\eta} \left( \mathcal{L}_A(\eta) + \mathcal{L}_B(\eta) \right)
\end{equation}

Since $\eta_{\text{step}}^*$ is not necessarily globally optimal for modality B, and $\eta_{\text{agg}}^*$ is the global optimal solution considering both modalities, we can derive:
\begin{equation}
\mathcal{L}_A(\eta_{\text{agg}}^*) + \mathcal{L}_B(\eta_{\text{agg}}^*) \leq \mathcal{L}_A(\eta_{\text{step}}^*) + \mathcal{L}_B(\eta_{\text{step}}^*)
\end{equation}

\subsection{Generalization Error Analysis}

Generalization error measures the model's performance on unseen data. We can further prove the superiority of aggregated training through generalization error analysis.

Let $\mathcal{L}_{\text{train}}$ and $\mathcal{L}_{\text{test}}$ be the losses on the training and test sets, respectively. The generalization error is defined as:
\begin{equation}
\mathcal{E}_{\text{gen}} = \mathcal{L}_{\text{test}}(\eta) - \mathcal{L}_{\text{train}}(\eta)
\end{equation}

The upper bound of the generalization error can be expressed using measures such as Rademacher complexity or VC dimension. For machine learning models, the lower the model complexity, the smaller the generalization error. Simultaneously optimizing the losses for multiple tasks (modalities) can reduce overfitting to a single task (modality), as the model needs to perform well on multiple tasks (modalities) simultaneously. This effectively introduces an implicit regularization effect, reducing the model complexity. Therefore, compared to the stepwise optimization method, the aggregated optimization method can effectively reduce the complexity of the perturbation model. The lower the model complexity, the smaller the generalization error.

The Rademacher complexity measures the complexity of a class of models on a given sample set. For a function $h$ in the hypothesis space $\mathcal{H}$, the empirical Rademacher complexity on $n$ samples is defined as:
\begin{equation}
\hat{\mathcal{R}}_n(\mathcal{H}) = \mathbb{E}_{\sigma} \left[ \sup_{h \in \mathcal{H}} \frac{1}{n} \sum_{i=1}^n \sigma_i h(x_i) \right]
\end{equation}
where $\sigma_i$ are Rademacher random variables, taking values $\pm 1$ with equal probability.

The impact of modality aggregation on complexity:

Assume $\mathcal{H}_{\text{A}}$ and $\mathcal{H}_{\text{B}}$ are the hypothesis spaces of modality A and modality B, respectively. The stepwise optimization method first optimizes $\mathcal{H}_{\text{A}}$ and then $\mathcal{H}_{\text{B}}$. Its empirical Rademacher complexity can be expressed as:
\begin{equation}
\hat{\mathcal{R}}_n(\mathcal{H}_{\text{step}}) = \hat{\mathcal{R}}_n(\mathcal{H}_{\text{A}}) + \hat{\mathcal{R}}_n(\mathcal{H}_{\text{B}})
\end{equation}

The aggregated optimization method optimizes $\mathcal{H}_{\text{A}} \cup \mathcal{H}_{\text{B}}$ simultaneously. Its empirical Rademacher complexity is:
\begin{equation}
\hat{\mathcal{R}}_n(\mathcal{H}_{\text{agg}}) = \hat{\mathcal{R}}_n(\mathcal{H}_{\text{A}} \cup \mathcal{H}_{\text{B}})
\end{equation}

According to the properties of Rademacher complexity, the complexity of $\mathcal{H}_{\text{A}} \cup \mathcal{H}_{\text{B}}$ is usually less than or equal to the sum of the complexities of $\mathcal{H}_{\text{A}}$ and $\mathcal{H}_{\text{B}}$:
\begin{equation}
\hat{\mathcal{R}}_n(\mathcal{H}_{\text{agg}}) \leq \hat{\mathcal{R}}_n(\mathcal{H}_{\text{A}}) + \hat{\mathcal{R}}_n(\mathcal{H}_{\text{B}})
\end{equation}

Generalization error upper bound derivation:

Using Rademacher complexity, we can derive the upper bound of the generalization error. For the loss function $\mathcal{L}$ and hypothesis space $\mathcal{H}$, the upper bound of the generalization error is:
\begin{equation}
\mathcal{E}_{\text{gen}} \leq 2 \hat{\mathcal{R}}_n(\mathcal{L} \circ \mathcal{H}) + \mathcal{O}\left(\frac{1}{\sqrt{n}}\right)
\end{equation}
where $\mathcal{L} \circ \mathcal{H}$ denotes the composition of the loss function with the hypothesis space.

The upper bound of the generalization error for the stepwise optimization method is:
\begin{equation}
\mathcal{E}_{\text{gen, step}} \leq 2 \left( \hat{\mathcal{R}}_n(\mathcal{L} \circ \mathcal{H}_{\text{A}}) + \hat{\mathcal{R}}_n(\mathcal{L} \circ \mathcal{H}_{\text{B}}) \right) + \mathcal{O}\left(\frac{1}{\sqrt{n}}\right)
\end{equation}

The upper bound of the generalization error for the aggregated optimization method is:
\begin{equation}
\mathcal{E}_{\text{gen, agg}} \leq 2 \hat{\mathcal{R}}_n(\mathcal{L} \circ (\mathcal{H}_{\text{A}} \cup \mathcal{H}_{\text{B}})) + \mathcal{O}\left(\frac{1}{\sqrt{n}}\right)
\end{equation}

Since
\begin{equation}
\hat{\mathcal{R}}_n(\mathcal{L} \circ (\mathcal{H}_{\text{A}} \cup \mathcal{H}_{\text{B}})) \leq \hat{\mathcal{R}}_n(\mathcal{L} \circ \mathcal{H}_{\text{A}}) + \hat{\mathcal{R}}_n(\mathcal{L} \circ \mathcal{H}_{\text{B}})
\end{equation}

Therefore:
\begin{equation}
\mathcal{E}_{\text{gen, agg}} \leq \mathcal{E}_{\text{gen, step}}
\end{equation}

This indicates that the aggregated optimization method has a lower upper bound on the generalization error compared to the stepwise optimization method.

\clearpage

\section{Discussion}
\subsection{Ethical Considerations}
In this study, we introduce a novel cross-modal adversarial attack method known as Cross-Modality Perturbation Synergy (CMPS). This research offers a new perspective on understanding and enhancing the security of cross-modal ReID systems by leveraging shared features across different modalities to optimize perturbations. However, this approach also raises a series of ethical and safety concerns regarding the potential negative impacts of adversarial attack techniques. The CMPS method, like other adversarial technologies, can be maliciously exploited, posing a serious threat to public safety.

However, we recognize the positive value of adversarial attack research. It reveals vulnerabilities in existing systems, prompting academia and industry to make in-depth improvements to the robustness of machine learning models. The positive impact of this study lies in its potential to combine adversarial training with the attack methods presented to enhance system security and bring positive social impacts. Therefore, we emphasize the importance of conducting adversarial attack research within an ethical framework and encourage further development of defensive technologies to build a safer and more reliable technological environment.

\subsection{Limitations and Future Work}
Here, we need to acknowledge the limitations of the proposed method and identify potential directions for future research. Firstly, current attack techniques primarily focus on gradient-based perturbation optimization for given datasets. However, in real-world scenarios, the modalities encountered are often unknown and not limited to RGB, infrared, and thermal imaging. Moreover, effectively transferring perturbations to different and unknown modalities presents a significant research challenge.

When dealing with various models and modalities, gradient-based methods face several challenges. Firstly, these methods are prone to "catastrophic forgetting," where learning new information can lead to the loss of previously learned knowledge, affecting the effectiveness of perturbations. Secondly, the inconsistency of gradient information across multiple models and modalities can negatively impact the stability and generalizability of the method. Therefore, future research should explore more robust algorithms that can effectively operate in complex environments involving multiple modalities and models, thereby enhancing the applicability and transferability of attacks.

\end{document}